\documentclass{article}

\usepackage{arxiv}

\usepackage[utf8]{inputenc} 
\usepackage[T1]{fontenc}    
\usepackage{hyperref}       
\usepackage{url}            
\usepackage{booktabs}       
\usepackage{amsfonts}       
\usepackage{nicefrac}       
\usepackage{microtype}      
\usepackage{lipsum}		
\usepackage{graphicx}
\usepackage{natbib}
\usepackage{doi}
\usepackage{amsmath}
\usepackage{longtable}
\usepackage{pdflscape}
\usepackage{multirow}
\usepackage{comment}
\usepackage{algorithm}
\usepackage{algpseudocode}

\title{The Traveling Thief Problem with Time Windows: Benchmarks and Heuristics}

\author{ \href{https://orcid.org/0009-0004-3572-249X}{\includegraphics[scale=0.06]{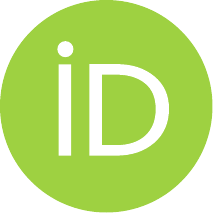}\hspace{1mm}Helen Yuliana Angmalisang}
        \\
        Optimisation and Logistics\\
	School of Computer Science and Information Technology\\
	Adelaide University\\
	Adelaide, SA 5000 \\
	\texttt{helen.angmalisang@adelaide.edu.au} \\
	\And
	\href{https://orcid.org/0000-0002-2721-3618}{\includegraphics[scale=0.06]{orcid.pdf}\hspace{1mm}Frank Neumann} \\
    Optimisation and Logistics\\
	School of Computer Science and Information Technology\\
	Adelaide University\\
	Adelaide, SA 5000 \\
	\texttt{frank.neumann@adelaide.edu.au} \\
}

\date{}

\renewcommand{\shorttitle}{The TTP with Time Windows: Benchmarks and Heuristics}

\hypersetup{
pdftitle={The Traveling Thief Problem with Time Windows: Benchmarks and Heuristics},
pdfsubject={The TTP with Time Windows: Benchmarks and Heuristics},
pdfauthor={Helen Yuliana Angmalisang, Frank Neumann},
pdfkeywords={Traveling Thief Problem, Time Windows, Heuristic Algorithm, Benchmark, Evolutionary Algorithm, Constraint},
}

\begin{document}
\maketitle

\begin{abstract}
While traditional optimization problems were often studied in isolation, many real-world problems today require interdependence among multiple optimization components. The traveling thief problem (TTP) is a multi-component problem that has been widely studied in the literature. In this paper, we introduce and investigate the TTP with time window constraints which provides a TTP variant highly relevant to real-world situations where good can only be collected at given time intervals. We examine adaptions of existing approaches for TTP and the Traveling Salesperson Problem (TSP) with time windows to this new problem and evaluate their performance. Furthermore, we provide a new heuristic approach for the TTP with time windows.
To evaluate algorithms for TTP with time windows, we introduce new TTP benchmark instances with time windows based on TTP instances existing in the literature. 
Our experimental investigations evaluate the different approaches and show that the newly designed algorithm outperforms the other approaches on a wide range of benchmark instances.
\end{abstract}


\section{Introduction}

There are many real-world applications that require joint routing and resource selection optimization, while being subject to time limitations and deadlines. 
Emergency services \cite{WoonKim2015,Pu2025}, home health care \cite{MARTINSQUEIROZ2025856}, and municipal waste management \cite{su16156585,HUANG201524} are some examples of such applications. In these problems, routing optimization and resource selection optimization cannot be addressed separately.
An important benchmark problem that shares the interacting characteristics of routing and resource selection optimization is the Traveling Thief Problem (TTP), which is an interdependent combination of the Traveling Salesman Problem (TSP) and the Knapsack Problem (KP). The TTP has gained significant attention in the evolutionary computation literature~\cite{Bonyadi2013,BonyadiMohammadReza2014Siaf,Polyakovskiy2014,Yafrani2016PopvsSing}. However, to the best of our knowledge, there is no study that investigates the time window constraint in TTP, which is highly relevant in many  real-world situations.

Nevertheless, with time window constraints, TTP becomes more complex as the travel speed in TTP depends on the distance and the current cumulative weight of the stolen items carried in the knapsack. Taking an item from a city may lead to either profit or loss, or cause late arrival at subsequent cities. However, the thief cannot make any profit if no item is taken. Moreover, with time windows, the feasible space becomes drastically narrower in a wide search space. Therefore, it is more challenging to find a feasible solution.

\subsection{Related Work}

\citet{Bonyadi2013} introduced and defined TTP as an optimization problem that is more similar to real-world problems for its combination and interdependence of the KP and TSP components. Since then, numerous studies have been conducted to find the best algorithms for solving TTP. Most of them used the Chained Lin-Kernighan (CLK) heuristic to create the initial tour \cite{MaityAlenrex2020Ehls,Zhang2021,Don2025,NguyenHoang2023SAwD,PathirageDonThilina2024TCCT}. The tour is then optimized using various algorithms. Operators like crossover and insertion are popular for optimizing the tour \cite{MeiYi2014IEoH,YiMei2015Hewg,VieiraDanielK.S.2017AGAf,AlharbiSaadT2018AHGA,NikfarjamAdel2022CDOf,Nikfarjam2022,NikfarjamAdel2022Edof}.

\cite{Faulkner2015ApproximateAT} proposed Pack and PackIterative algorithms as item scoring strategies for the packing plan. They and some studies afterwards \cite{WagnerMarkus2018Acso,ELYAFRANI2018231,Zhang2021,NguyenHoang2023SAwD} found that repetition of CLK and PackIterative algorithm (S5) shows particularly strong results. Compared to population-based algorithms, S5 which is a single-solution algorithm also performs better on the TTP \cite{Yafrani2016PopvsSing}. Moreover, it is found that the initial tour really affects the final objective value \cite{Faulkner2015ApproximateAT}. BitFlip, Insertion, Pack, and PackIterative have become popular operators that have been used in many investigations afterwards \cite{VieiraDanielK.S.2017AGAf,AlharbiSaadT2018AHGA,Alharbi2018TheDA,MaityAlenrex2020Ehls,Zhang2021,ChagasJonatasB.C.2022Awmf}. 

\cite{WuijtsRogierHans2019Iott} found that the TTP fitness landscape contains numerous local optima, while the global optimum lies within a narrow region of attraction. Therefore, it is suggested to increase the size of the perturbations to improve the chances of discovering the global optimum and to improve the accuracy of fitness estimates. These findings underlie the development of our proposed algorithms.  

Regarding time windows, there have been many studies in decades on the Traveling Salesman Problem with Time Windows (TSPTW), which is a component of TTPTW. \cite{helsgaun2017extension} improved the classic Lin-Kernighan algorithm \cite{lin1973effective}, for constrained TSP and vehicle routing problems (VRP), named LKH-3. The LKH-3 algorithm can solve different variants of those problems, including TSPTW. Subsequently, \cite{Zheng_He_Zhou_Jin_Li_2021} proposed a Variable Strategy Reinforced LKH (VSR-LKH) and VSR-LKH-3 algorithms and combined them with three reinforcement learning methods, namely Q-learning, Sarsa, and Monte Carlo, for the TSP. VSR-LKH-3 was reported to have better performance in solving TSPTW compared to LKH-3. However, to the best of our knowledge, no study or algorithm has been specifically proposed for TTP with time windows. 

\subsection{Our Contribution}

This study extends the Traveling Thief Problem with time window constraints (TTPTW). The time windows constraint adds a condition in which the thief has to wait if he is early. This condition changes the objective function, as the total duration of the trip may increase and affect the rental fee.

For the evaluation, we construct the benchmark with varying level of time window tightness. Furthermore, we propose a set of evolutionary algorithms, named Dual Search Evolutionary Algorithm (DSEA), that utilize a new tour initialization algorithm, modified operators, and different packing plan repair approaches. We compared the performance of the packing plan repairs and analyzed them to study which repair works best in this problem. The effect of using tour repair is also investigated by comparing the variants of the DSEA algorithm. 

For comparison, we adapt the state-of-the-art TTP and TSPTW algorithms 
and evaluate their performance on TTPTW instances. We also implement our new tour initialization algorithm to 
the TTP heuristics and evaluate their performance on TTPTW. We compare and analyze the effect of using the tour initialization algorithm and the variation of operators on these TTP optimizers. Finally, we compare the performance of DSEA, S4, S5, C5, LKH-3, and VSR-LKH-3 algorithms.

The rest of the paper is structured as follows. In Section \ref{TTP}, we provide an overview of general TTP. Section \ref{TTPTW} introduces the Traveling Thief Problem with Time Windows. Section \ref{CurrentApproaches} explains how we adapt state-of-the-art TTP and TSPTW algorithms for TTPTW. Section \ref{Proposed} describes the proposed algorithms, their components, and the difference among the variants. Section \ref{benchmark} explains the benchmark problem used in this study. Section \ref{ExpInv} presents the experimental settings and results. Finally, Section \ref{concl} provides the conclusion.

\section{Traveling Thief Problem}\label{TTP}
Given are a set of cities \(N=\{1,2,\dots,n\}\) and a set of items \(M=\{1,2,\dots,m\}\). The items are distributed among the cities, except city 1, and each item $j \in M$ has a profit $p_j$ and a weight $w_j$. The distance between any pair of cities $i,j \in N$, $d_{ij}$, is known. The thief has to start from city 1 and return there after visiting all the other cities once. The thief can steal some items as long as the total weight of the stolen items does not exceed the knapsack capacity $W$. The knapsack rental rate $R$ is charged for each time unit used during the tour. The minimal and maximal speed of the thief, $v_{min}$ and $v_{max}$, are known. The objective of this problem is to find a tour $\Pi = (x_1, x_2, ..., x_n), x_i \in N$ and a packing plan $P = (y_1, y_2, ..., y_m) \in \{0,1\}^m $ to maximize profit. The formulation of the objective functions is as follows.

\begin{equation} \label{obj_func_ttp}
    \begin{aligned}
    \max_{\Pi,\,P} \quad & Z (\Pi,P)= \sum_{j=1}^{m} p_j y_j - R \left( \frac{d_{x_n x_1}}{v_{max}-vW_{x_n}} + \sum_{i=1}^{n-1} \frac{d_{x_i x_{i+1}}}{v_{max} - v W_{x_i}} \right)\\
    \end{aligned}
\end{equation}

where
 $ v=\frac{v_{max}-v_{min}}{W}.$  
If $y_j = 0$, then the item $j \in M$ is not stolen. Otherwise, it is stolen. $W_x$ denotes the total weight of the knapsack when the thief leaves the city $x \in N$.

\section{Traveling Thief Problem with Time Windows}\label{TTPTW}

In TTPTW, the city can only be visited in certain time windows. Given a set of time windows in each city $T=\{q_1,\dots,q_n \}$, where each $q_i=(L_i,U_i)$. $L_i$ and $U_i$ are the lower and upper bound of the time window in city $i \in N$, respectively. The planned arrival time in city $x_i$ is given as

\begin{equation} \label{PlannedArrivalTime}
    t^*_{x_i} (\Pi,P)= t_{x_{i-1}} (\Pi,P)+\frac{d_{x_{i-1}x_i}}{v_{max}-vW_{x_{i-1}}}.
\end{equation}

where $ v=\frac{v_{max}-v_{min}}{W}$ and represents a component of the travel speed update.
We have $t_1 = t^*_1 = 0$ and $t_{x_i}$ is the exact arrival time in city $x_i$ given by

\begin{equation} \label{t_{x_i=L_i}}
    t_{x_i}(\Pi,P) = \max(t^*_{x_i}(\Pi,P),L_{x_i}).
\end{equation}

The planned and exact arrival time may be different as the thief has to wait during the trip if the planned arrival time $t^*_{x_i}$ is earlier than the lower bound of the time window in the city  ($t^*_{x_i} < L_{x_i})$. If the thief has to wait, the waiting duration may affect the arrival times in the subsequent cities also. Therefore, the formulation of objective function and constraints for TTPTW is the following:

\begin{equation} \label{obj_func_ttptw}
\begin{aligned}
\max_{\Pi,\,P} \quad & Z (\Pi,P) = \sum_{j=1}^{m} p_j y_j - R \left( \frac{d_{x_n x_1}}{v_{\max}-vW_{x_n}} + t_{x_n} \right) \\
\text{s.t.} \quad & 
    \sum_{i=1}^{n} \sum_{j=1}^{m} w_j y_{ij} \leq W \\
    \quad & x_1 = x_{n+1} = 1 \\
    \quad & x_a \ne x_b, \quad \forall x_a, x_b \in N\setminus\{1\} \\
    \quad & t_{x_i} \leq U_{x_i}
\end{aligned}
\end{equation}

All constraints are deterministic.
In order to handle infeasibility due to time window constraints, the value of time window constraints violation ($cv$) is:

\begin{equation} \label{cv}
    cv(\Pi, P) = \sum_{i=1}^{n} max(0,t_{x_i}(\Pi, P)-U_{x_i})
\end{equation}

This study uses the constraint-objective value separation technique \cite{DEB2000311}, where the violations of the time-window constraints are calculated separately and not combined with the objective function value. A solution with a lower constraint violation is considered better. If two solutions have the same constraint violation value, then the one with the higher objective function value is considered better.

We do not employ a penalty-based method due to the highly variable scale of the objective function, which can take extremely negative values due to prolonged travel times even for feasible solutions. Under such condition, the penalty-based method may fail to properly balance constraint violation and objective value. Since this study considers hard time windows, it is very important to ensure that infeasible solutions are not preferred over feasible ones regardless of their objective values.

\section{Adaptation of Existing Approaches}\label{CurrentApproaches}

At first, we adapt the state-of-the-art TTP and TSPTW  heuristics to solve TTPTW.

\subsection{Adaptation of TTP Heuristics}
We adapt S4, S5, C5 algorithms \cite{Faulkner2015ApproximateAT} to TTPTW by adding the time windows and the procedure to calculate the violation of the time window constraints in the evaluation function. We also modified all the solution comparison procedures using the constraint-objective value separation technique \cite{DEB2000311}. In the preliminary study using the same experimental setup in Section \ref{ExpInv}, we found that these algorithms cannot find any feasible solution in all datasets, except S4 in 51-A with positive $l$, as shown in Table \ref{tab:S4S5}.

In this study, we applied our proposed tour initialization algorithm (Section \ref{tourinit}) to these algorithms (S4-Initial, S5-Initial, and C5-Initial) to create the initial tour at the beginning of the algorithms.

\subsection{Adaptation of TSPTW Heuristics}
We also adapt LKH-3 and VSR-LKH-3 for TTPTW, because they are state-of-the-art algorithms for TSP, including TSPTW. After finding the tour for TSPTW, the process continues to generate the packing plan based on the tour using PackIterative \cite{Faulkner2015ApproximateAT}. With the same experimental setup as in Section \ref{ExpInv}, we conducted the evaluation of these algorithms. Based on the results as shown in Table \ref{tab:S4S5}, VSR-LKH-3 barely finds a feasible solution, while LKH-3 can find feasible solutions, but not in all runs in the instances with more than 100 cities. The feasible rate decreases as the instance gets bigger.

\begin{table}[t]
\small
\centering
  \caption{Feasible Rate of S4, S5, C5, LKH-3, and VSR-LKH-3 on TTPTW Benchmark}
  \label{tab:S4S5}
  \begin{tabular}{ccccccc}
    \toprule
    Instance & $l$ & S4 & S5 & C5 & LKH-3 & VSR-LKH-3 \\
    \midrule
    \multirow{4}{*}{51-A} & 100 & $40\%$ & $0\%$ & $0\%$ & $100\%$ & $13\%$\\
    & 1000 & $100\%$ & $0\%$ & $0\%$ & $100\%$ & $50\%$\\
    & -100 & $0\%$ & $0\%$ & $0\%$ & $100\%$ & $0\%$\\
    & -1000 & $0\%$ & $0\%$ & $0\%$ & $0\%$ & $0\%$\\
    \multirow{4}{*}{100-A} & 100 & $0\%$ & $0\%$ & $0\%$ & $97\%$ & $0\%$\\
    & 1000 & $0\%$ & $0\%$ & $0\%$& $100\%$ & $0\%$\\
    & -100 & $0\%$ & $0\%$ & $0\%$& $97\%$ & $0\%$\\
    & -1000 & $0\%$ & $0\%$ & $0\%$& $97\%$ & $0\%$\\
    \multirow{4}{*}{150-A} & 100 &  $0\%$ & $0\%$ & $0\%$ & $80\%$ & $0\%$\\
    & 1000 & $0\%$ & $0\%$ & $0\%$ & $97\%$ & $0\%$\\
    & -100 & $0\%$ & $0\%$ & $0\%$ & $40\%$ & $0\%$\\
    & -1000 & $0\%$ & $0\%$ & $0\%$ & $17\%$ & $0\%$\\
    \multirow{4}{*}{225-A} & 100 &  $0\%$ & $0\%$ & $0\%$ & $67\%$ & $0\%$\\
    & 1000 & $0\%$ & $0\%$ & $0\%$ & $0\%$ & $0\%$\\
    & -100 & $0\%$ & $0\%$ & $0\%$ & $0\%$ & $0\%$\\
    & -1000 & $0\%$ & $0\%$ & $0\%$ & $0\%$ & $0\%$\\    
    \bottomrule
  \end{tabular}
\end{table}

\section{Dual Search Evolutionary Algorithm (DSEA)}\label{Proposed}

Because the adapted TTP and TSPTW algorithms do not perform well in solving TTPTW, we develop a set of algorithms for TTPTW, named Dual-Search Evolutionary
Algorithm (DSEA). In this section, we discuss the main algorithm of DSEA, its variants, their components, and the packing plan repair approaches employed in this study.

\subsection{Outline of the Algorithm and Its Variants}

The pseudo-code of Dual Search Evolutionary Algorithm (DSEA) is shown in Algorithm \ref{alg:newAlg}. It begins with Tour Initialization (Section \ref{tourinit}) to generate an initial tour  (line 1). The process then continues to initialize the packing plan based on the tour using PackIterative algorithm \cite{Faulkner2015ApproximateAT} (line 2). The generated tour and packing plan are saved as the initial best solution (line 3). The current best solution is then optimized using Operator 1 (O1) and Operator 2 (O2) (line 10). After O2 ends, the best solution is updated if the new solution is better than the current best one (lines 11-15). Subsequently, a new packing plan is created using PackIterative based on the new tour (line 16). If the new tour and newly generated packing plan are better than the current best solution, the best solution is updated (lines 16-18). If in $\rho$ consecutive main loops, no better solution is found, the tour mutation is conducted to enhance
exploration before the tour is optimized again (lines 7-9). Mutation, O1, O2, PackIterative, and best solution update are repeated until the stopping criterion.

The mutation procedure is only a simple swap of two random cities in the current best tour for a random number of times between 1 and $h$. There is also an adaptation for the Pack algorithm by \cite{Faulkner2015ApproximateAT}. In this study, the algorithm performs an objective function and a constraint violation evaluation every time it tries to add an item to the knapsack. Our preliminary study shows that this setting generates a significantly better packing plan. In this study, three variants of DSEA are investigated in this study. Each of them has a different packing plan repair approach. Table \ref{tab:DSEAvar} shows the differences between the DSEA variants.

\begin{table}[t]
\small
\centering
  \caption{Variants of DSEA}
  \label{tab:DSEAvar}
  \begin{tabular}{cccc}
    \toprule
    Variant & O1 & O2 & Packing Plan Repair\\
    \midrule
    DSEA$_1$ & Topo & Rain & No Repair\\
    DSEA$_2$ & Topo & Rain & Repack\\
    DSEA$_3$ & ITP & IIP & Integration of Repair
and Optimizer\\
    \bottomrule
  \end{tabular}
\end{table}

\begin{algorithm}[t]
    \small
    \caption{Dual Search Evolutonary Algorithm (DSEA)}\label{alg:newAlg}
    \begin{algorithmic}[1]
        \State $\Pi_{B} \gets$ TourInitialization()
        \State $P_{B} \gets$ PackIterative$(\Pi_{B}, c, \delta, q, e)$
        \State $B \gets (\Pi_{B},P_{B})$
        \State $k \gets 0$
        \While {not meet stopping criterion}
            \State $\Pi_x \gets \Pi_{B}$
            \If{$k > \rho$}
                \State $\Pi_x \gets$ Mutate($\Pi_{x}$)
            \EndIf
            \State $(\Pi_{x}, P_{x}) \gets$ O2(O1$(\Pi_{x},P_{B}))$
            \State $X_1 \gets (\Pi_x,P_{x})$
            \If {$X_1$ is better than $B$}
                \State $B \gets X_1$
                \State $k = 0$    
            \EndIf
            \State $X_2 \gets (\Pi_x,$ PackIterative$(\Pi_x, c, \delta, q, e))$
            \If {$X_2$ is better than $B$}
                \State $X_1 \gets X_2$
                \State $k = 0$
            \EndIf

            \State $k += 1$
        \EndWhile
    \end{algorithmic}
\end{algorithm}



\subsection{Tour Initialization}\label{tourinit}

As \cite{Faulkner2015ApproximateAT} stated that the initial tour really affects the final objective value, tour initialization is an important part that can help the main algorithm in finding a feasible solution from the beginning. Instead of using CLK like majority studies on TTP, we proposed a tour initialization algorithm. This is because the shortest route, which CLK tries to find, is not always feasible due to time-window constraints. Moreover, the time windows significantly reduce the feasible area of a very large search space. It makes the search for feasible tours using CLK computationally very expensive. 
The tour initialization adapts the nearest neighborhood approach by adding memory and scoring procedure to generate a tour. The scoring procedure calculates the arrival time from the previous city $x_i$ to all unvisited cities $j \in N$.

\[
score_j (\Pi,P)=
\begin{cases}
travelTime_{x_i j}, & {L_j \leq t^*_j \leq U_j}, \\
-(travelTime_{x_i j}+M(t^*_j-U_{x_i})), & \text{otherwise}
\end{cases}
\]
where $travelTime_{x_i j} = \frac{d_{x_i j}}{v_{max}}$ and $M$ is a large positive value to distinguish between the cities predicted to be reached on time, early, or late. If the arrival time in the next city, $x_{j}$, satisfies the time window in the city ($L_j \leq t^*_j \leq U_j$), the score of that city is the travel time from $x_i$ to $j$. In contrast, if the time window is violated, then the score will be penalized. If the thief is late, the score will be negative, so it can be prioritized, whereas if the thief is too early, the city will be considered less important to visit next. This scoring will create a sequence of unvisited cities.

When a city is predicted to be visited late, the algorithm does not attempt to explore alternative sequences. It only avoids the constraint violation from increasing. This approach minimizes the computational effort as an initialization algorithm, because the solution will be refined in the optimization process later. It also avoids unnecessary searching when the instance has no any feasible solution.

It starts in city $1$ and the scoring is conducted. The cities, except city $1$, are then sorted according to their scores. This sequence and the reverse one are memorized as the first and second tours. After that, the city with the lowest score will be visited next. For example, if city $4$ got the lowest score, then the visited cities are $(1, 4)$, while city $2, 3, 5, 6, \ldots$ will be scored again from city $4$. After all the cities are visited, a new tour is created based on the visit sequence and is memorized as the third tour. Its reverse is also memorized as the fourth one. The tour with the least constraint violation and the highest objective function value among all memorized tours is selected as the result.

\subsection{Search Operators}

The DSEA uses two different search operators in the same run (line 10 in Algorithm \ref{alg:newAlg}). One operator is 2-opt-based, while the other is insertion-based, as our preliminary study shows that
the combination of these two operators helps the algorithm perform better and both of them complement each other.

For the 2-opt-based operator, we proposed Two-Opt with Perturbation Operator (Topo) 
and Integration of Two-opt and Pack (ITP). 
DSEA uses 2-opt as the basic move instead of the k-opt operator like Lin-Kernighan and LKH-3 for efficiency reasons and because \cite{Liu2021ComparativeOperators} show that the 2-opt operator makes the most improvement compared to the others, e.g, 3-opt and 4-opt. For the insertion-based operator, we proposed Random Insertion (Rain) and Integration of Insertion and Pack (IIP). Insertion is used as the basic move, 
because S4 \cite{Faulkner2015ApproximateAT} which uses only insertion outperforms S5 and C5 according to Table \ref{tab:S4S5} and Table \ref{tab:S4S5Init}. 

Topo is used in DSEA$_{1}$ and DSEA$_{2}$ to optimize the current tour. It tries to reverse the subtour at random. After reverse, the new tour is evaluated with a probability of $\mu$. If the new tour is better than the current tour, the current tour is updated with the new one. Otherwise, there will be packing plan repair if the algorithm uses the repair procedure.

It takes the probability of $\mu$ as the perturbation to prevent the algorithm from being trapped in local optima. If not evaluated, the algorithm continues to reverse the next subtour. In other words, the probability of $\mu$ provides the possibility of creating a different tour, rather than conducting an evaluation after every reverse to enhance the exploration. This process repeats $\theta$ times.

Random Insertion (Rain) is used in DSEA$_{1}$ and DSEA$_{2}$ to optimize the tour by randomly inserting a city into another index in the tour. If the insertion gives a better result, the current tour will be updated with the new one. Otherwise, there will be packing plan repair if the algorithm uses the repair scheme. The operator also repeats $\theta$ times. Randomness is used to give different generated tours that might be a better tour, while without randomness, the insertion operator might give the same tour because it always does insertion in the exactly same procedure in each iteration.

Integration of Two-opt and Pack (ITP) is only used in DSEA$_3$. In the beginning, the item scores of all items are calculated. Then, the tour is generated by 2-opt randomly. After that, the item scores in the affected cities are updated. If a random number is greater than $\mu$, the newly found solution will be evaluated and the current best solution updated. When the repetition counter is divisible by $\rho$, the packing plan is updated using $Pack$; otherwise, the evaluated packing plan is the current one. If the new solution is better than the current best one, the best solution is updated. Like $Topo$, this entire procedure is repeated $\theta$ times.

Integration of Insertion and Pack (IIP) is only used in DSEA$_3$. It is similar to ITP, but the new tour is generated using random insertion and only the items in the inserted city are updated. We do not update the items in the rest city that move slightly because of the insertion to make it efficient. In addition, the evaluation is performed every time there is a new tour.

 

\subsection{Packing Plan Repair}

For packing plan repair, we propose two approaches, namely Repack and Integration of Repair and Optimizer. Repack approach is only used in DSEA$_2$. It is a modification of Pack algorithm, designed specially for time windows problem. The modification is based on the understanding that in TTP there are items that are not worth taking as it can make a loss, because the profit is too small, but the weight is too heavy.

Firstly, it calculates the contributions of each item that was taken before the thief was late. If the solution after deleting the item is better than taking it, the item is deleted. After that, we use the similar pack algorithm, but only for the items that are previously not in the list to be taken, for the additional items until the knapsack is full. This repair approach is conducted if the counter in $Topo$ or $Rain$ is divisible by $\beta$.

Meanwhile, integration of repair and optimizer approach is only used in DSEA$_3$. It is proposed based on the consideration that after Topo or Rain operator, the tour changes, but only in some part. ITP and IIP fall under this approach. Instead of using PackIterative every time a new tour is generated, we use Pack algorithm as in the preliminary study, we found that Pack and PackIterative algorithms mostly give the same packing plan. Moreover, Pack is much faster than PackIterative.


Similarly to the Repack approach, we do not repair the packing plan every time in order to save time. Thus, after updating the item scores, if the counter is divisible by $\beta$, the packing plan is repaired using Pack.


\section{Benchmark Instances for the the Traveling Thief Problem with Time Windows}\label{benchmark}

To investigate the performance of the proposed algorithm in solving the Traveling Thief Problem with Time Windows, we conducted an algorithm evaluation. In the following, we describe the new benchmark instances that we generated for TTPTW.



We use classical TTP benchmark instances introduced in~\cite{Polyakovskiy2014}\footnote{Instances available at \url{https://cs.adelaide.edu.au/~optlog/CEC2014Comp/}} and add time windows to obtain TTPTW instances. The time windows are generated using the same method as in \cite{DumasTSPTW1995} and \cite{Bi2024}, specifically following the hard setting.
The lower bound of the time windows in each city $i$, $L_i \sim Uniform(max(0,t-l), t)$ and the upper bound $U_i \sim Uniform(t', (t'+l))$, where $\forall l \in \{100, 1000, -100, -1000\}; t_i$ and $t'_i$ denote the arrival time of the thief in each city $i$, traveling at maximum speed $v_{max}$ and minimum speed $v_{min}$, respectively. Note that $l$ denotes a parameter controlling the tightness of the time windows and does not represent the actual width of the time window. Considering that there are negative $l$, to ensure that there is a feasible solution that satisfies all the time windows constraints, we modified them into the more general formula: $L_i = max(0, min(t_i-random(0,l),t'_i)))$ and $U_i = max(t'_i+random(0,l),t_i)$. To ensure that the time windows in an instance with smaller $l$ is smaller than the greater one, we use more conditional that if $l_a > l_b$, $L^{(l_a)}_i \leq L^{(l_b)}_i$ and $U^{(l_a)}_i \geq U^{(l_b)}_i$ where $L^{(l_c)}_i$ and $U^{(l_c)}_i$ denote the lower bound and upper bound of the time window in city $i$ in the same TTP instance where $\exists l_c \in \{100, 1000, -100, -1000\}$.

We used various $l$ values to study the performance of algorithms in different time window widths. We also consider  very tight time windows as it naturally arises in several real-world applications such as emergency response services \cite{Pu2025} and food delivery \cite{su17062771}. It also serves for the robustness evaluation of the proposed algorithms.

There are two types of instances: Type A which uses the optimal route in the corresponding TSP instance for generating the time windows, and the instances Type B which use the other tour than the TSP optimal tour to generate the time windows. Instances type A are used for the result comparison to the basic TTP as the optimal solution in the instances are already provided. Thus, the performance of the algorithms is measurable, as it contains the optimal solution in the basic TTP for instances with positive $l$.

Meanwhile, Type B is created as in the real-world, the time windows pattern is not always in line with the shortest route. The TSP optimal tour is not always the optimal tour in TTP, including in the problem with time windows. The tour for generating the time windows is generated randomly.

The instances consist of various numbers of cities, ranging from $51$ to $1000$, with different numbers of items, while the number of items ranges from $50$ to $999$. In addition, we also examined different knapsack instance types. $x$-u denotes the instances where profit and weight are uncorrelated, while $x$-usw denotes the instance that has similar weights for all items but different profits \cite{Polyakovskiy2014}. $x$-A, $x$-B, and $x$-2 instances have bounded strongly correlated settings from \cite{Polyakovskiy2014}, $x$-2 instances have the type A time windows and larger knapsack capacity and renting ratio ($R$) than the $x$-A and $x$-B. Meanwhile, $x$-A and $x$-B share the same knapsack component, knapsack capacity and $R$, but have different time windows. Table \ref{tab:instancelist} presents the instances, along with the aliases assigned to each for ease of reference.

\begin{table}[t]
\small
\centering
  \caption{List of TTP Instances Used in This Paper}
  \label{tab:instancelist}
  \begin{tabular}{lcc}
    \toprule
    Original Instance Name & Type & Alias \\
    \midrule
    eil51\_n50\_bounded-strongly-corr\_01.ttp & A & 51-A \\
    eil51\_n50\_bounded-strongly-corr\_02.ttp & A & 51-2 \\
    eil51\_n50\_uncorr\_01.ttp & A & 51-u \\
    eil51\_n50\_uncorr-similar-weights\_01.ttp & A & 51-usw \\
    eil51\_n50\_bounded-strongly-corr\_01.ttp & B & 51-B \\
    kroC100\_n99\_bounded-strongly-corr\_01.ttp & A & 100-A \\
    kroC100\_n99\_bounded-strongly-corr\_01.ttp & B & 100-B \\
    ch150\_n149\_bounded-strongly-corr\_01.ttp & A & 150-A \\
    ch150\_n149\_bounded-strongly-corr\_02.ttp & A & 150-2 \\
    ch150\_n149\_uncorr\_01.ttp & A & 150-u \\
    ch150\_n149\_uncorr-similar-weights\_01.ttp & A & 150-usw \\
    ch150\_n149\_bounded-strongly-corr\_01.ttp & B & 150-B \\
    tsp225\_n224\_bounded-strongly-corr\_01.ttp & A & 225-A \\
    tsp225\_n224\_bounded-strongly-corr\_02.ttp & A & 225-2 \\
    tsp225\_n224\_uncorr\_01.ttp & A & 225-u \\
    tsp225\_n224\_uncorr-similar-weights\_01.ttp & A & 225-usw \\
    tsp225\_n224\_bounded-strongly-corr\_01.ttp & B & 225-B \\
    rat575\_n574\_bounded-strongly-corr\_01.ttp & A & 575-A \\
    rat575\_n574\_bounded-strongly-corr\_01.ttp & B & 575-B \\
    dsj1000\_n999\_bounded-strongly-corr\_01.ttp & A & 1000-A \\
    dsj1000\_n999\_bounded-strongly-corr\_01.ttp & B & 1000-B \\
    \bottomrule
  \end{tabular}
\end{table}

\section{Experimental investigations}\label{ExpInv}
We implemented the proposed algorithms, DSEA, using Python language. For  comparison, we use the original code of S4, S5, C5, LKH-3, and VSR-LKH-3~\footnote{available at \url{http://webhotel4.ruc.dk/~keld/research/LKH-3/} and VSR-LKH-3 for TSPTW in \url{https://github.com/JHL-HUST/VSR-LKH-V2}}. S4, S5, and C5 were written in Java, whereas LKH-3 and VSR-LKH-3 were written in C. At the end, the final objective and constraint violation values were calculated using the same evaluation function code to calibrate the results. For LKH-3 and VSR-LKH-3, after the final tour is generated, we call $PackIterative$ to find the packing plan based on the tour using the Python language. We used the custom parameters reported in \cite{helsgaun2017extension} and \cite{ZHENG2023110144}. For DSEA, we set $\mu = 0.5, \theta = 1000, \beta = 10, \rho = 5, M = 10000, h = 10$ which were selected based on  preliminary experiments that several candidate values were evaluated and those that provided the best performance were chosen, and $c = 5, \delta = 2.5, q = 20, e = 0.1$ as suggested in \cite{Faulkner2015ApproximateAT}.

All algorithms are evaluated in 30 independent runs, where in each run, all algorithms stop after 1,000,000 Function Evaluations (FE). We do not use CPU time as the stopping criterion, because the algorithms are not written in the same programming language. According to \cite{RAVBER2022109478}, when CPU time cannot be applied, max FE is the best stopping criterion. We run all algorithms on an Ubuntu system with an Intel(R) Core(TM) i7-1255U
(1.70 GHz CPU, 10 cores) and 16 GB of RAM.  The results were analyzed using the Kruskal-Wallis test
for the significance test and The Dunn-Bonferroni post-hoc test for
the pair-wise test. 

\subsection{Impact of Packing Plan Repair Mechanisms}

The evaluation results of the proposed algorithms with different packing plan mechanisms, DSEA$_1$-DSEA$_3$, are presented in Table \ref{tab:ResultdifferentPackRepair}. mean OB, std OB, FR, and stat denote the mean of the objective function values,
the standard deviation of the objective function values, feasible rate, and stat denotes the statistical test results, respectively. For example, $2(+), 3(-)$ means that DSEA$_1$ is significantly better than DSEA$_2$ and significantly worse than DSEA$_3$. $1(*), 3(-)$ means that there is no significant difference between DSEA$_1$ and DSEA$_2$, while DSEA$_2$ is significantly worse than DSEA$_3$. The Feasible Rate (FR) is the ratio of the number of runs that find a feasible solution to the total number of runs. 

To provide a clearer overall comparison, we rank the algorithms according to their mean values of the objective function for each instance. The result is that DSEA$_1$, DSEA$_2$, and DSEA$_3$ have an average rank of 1.42, 2.33, and 2.00, respectively. In other words, DSEA$_1$, the proposed algorithm without packing plan repair, outperforms the other variants that use the packing plan repair mechanism, DSEA$_2$ and DSEA$_3$. Furthermore, although  all DSEA variants have the same feasibility rate in almost all instances, there is a difference in 225-A with $l = -1000$. DSEA$_1$ consistently finds feasible solutions, whereas the other variants fail in some runs. This is because the repair procedure requires an additional computation cost. Taking into account that the difference between DSEA$_1$ and the others is only the use of the packing plan repair, it can be inferred that the absence of the repair mechanism benefits the algorithm. Rather than focusing on repair, DSEA$_1$ explores more tours and generates the packing plan only in the last part of the main loop. 

However, DSEA$_1$ is stuck in the local optima in 1000-A and 1000-B instances due to the absence of packing plan repair. The proposed algorithm that employs the integration of the repair and optimizer, DSEA$_3$, on the other hand, shows better results in these instances because it explores better packing plans. DSEA$_3$ also shows a significantly better result in 51-B and not worse than the others in 51-A. It shows that DSEA$_3$ has good performance in small bounded-strongly-correlated instances only, because in the other knapsack categories (uncorrelated and uncorrelated-similar-weights), it is outperformed by DSEA$_1$.  

Meanwhile, DSEA$_2$, the proposed algorithm that uses Repack, barely obtains a better result, showing its inferior performance. This is because Repack is quite expensive. The results also show that Repack is less accurate than the integration of repair and optimizer approach (DSEA$_3$) in repairing the packing plan.

\begin{table}[t]
    \tiny
    \centering
  \caption{Comparison on Different Packing Plan Repair Algorithms}
  \label{tab:ResultdifferentPackRepair}
  \begin{tabular}{ccrrccrrccrrcc}
    \toprule
    \multirow{2}{*}{Instance} & \multirow{2}{*}{$l$} & \multicolumn{4}{c}{DSEA$_1$ (1)} & \multicolumn{4}{c}{DSEA$_2$ (2)} & \multicolumn{4}{c}{DSEA$_3$ (3)} \\ \cline{3-14}
    & & mean OB & std OB & FR & stat & mean OB & std OB & FR & stat & mean OB & std OB & FR & stat \\
    \midrule
    \multirow{4}{*}{51-A} & 100 & 3834.78 & 28.99 & $100\%$ & 2(*),3(*) & 3836.13 & 36.09 & $100\%$ & 1(*),3(*) & 3847.39 & 57.81 & $100\%$ & 1(*),2(*)\\
    & 1000 & 3749.81 & 199.21 & $100\%$ & 2(*),3(-) & 3715.53 & 208.95 & $100\%$ & 1(*),3(-) & 3936.83 & 156.35 & $100\%$ & 1(+),2(+)\\  
    & -100 & 3468.07 & 0 & $100\%$ & 2(+),3(*) & 3467.91 & 0.20 & $100\%$ & 1(-),3(-) & 3524.47 & 87.85 & $100\%$ & 1(*),2(+)\\
    & -1000 & -581.81 & 78.99 & $0\%$ & 2(*),3(*) & -607.15 & 34.79 & $0\%$ & 1(*),3(*) & -645.55 & 111.49 & $0\%$ & 1(*),2(*)\\
    \multirow{4}{*}{100-A} & 100 & 4533.98 & 0 & $100\%$ & 2(*),3(*) & 4533.98 & 0 & $100\%$ & 1(*),3(*) & 4533.98 & 0 & $100\%$ & 1(*),2(*) \\
    & 1000 & 4533.98 & 0 & $100\%$ & 2(*),3(+) & 4533.98 & 0 & $100\%$ &  1(*),3(+) & 4531.47 & 9.45 & $100\%$ & 1(-),2(-) \\ 
    & -100 & 4513.85 & 0 & $100\%$ & 2(*),3(*) & 4513.85 & 0 & $100\%$ & 1(*),3(*) & 4513.85 & 0 & $100\%$ & 1(*),2(*)\\
    & -1000 & 4246.19 & 0 & $100\%$ & 2(*),3(*) & 4245.44 & 2.86 & $100\%$ & 1(*),3(*) & 4244.82 & 4.33 & $100\%$ & 1(*),2(*)\\
    \multirow{4}{*}{150-A} & 100 & 8772.66 & 12.16 & $100\%$ & 2(+),3(*) & 8736.63 & 63.04 & $100\%$ & 1(-),3(*) & 8755.52 & 32.82 & $100\%$ & 1(*),2(*) \\
    & 1000 & 8732.84 & 63.46 & $100\%$ & 2(+),3(+) & 8681.79 & 82.66 & $100\%$ & 1(-),3(+) & 8595.41 & 130.16 & $100\%$ & 1(-),2(-)  \\ 
    & -100 & 8646.24 & 12.16 & $100\%$ & 2(+),3(+)  & 8619.09 & 53.75 & $100\%$ & 1(-),3(*)  & 8619.19 & 49.44 & $100\%$ & 1(-),2(*) \\
    & -1000 & 7487.09 & 58.05 & $100\%$ & 2(+),3(+)  & 7409.64 & 87.64 & $100\%$ & 1(-),3(*)   & 7411.04 & 69.37 & $100\%$ & 1(-),2(*) \\    
    \multirow{4}{*}{225-A} & 100 & 14319.98 & 333.14 & $100\%$ & 2(+),3(+)  & 14049.71 & 145.76 & $100\%$ & 1(-),3(*) & 14040.05 & 176.09 & $100\%$ &  1(-),2(*)\\
    & 1000 & 11916.29 & 989.30 & $100\%$ & 2(+),3(*) & 9784.18 & 707.95 & $100\%$ & 1(-),3(-) & 11886.63 & 1080.67 & $100\%$ & 1(*),3(+) \\ 
    & -100 & 14822.83 & 11.48 & $100\%$ & 2(+),3(+)  & 14788.32 & 16.47 & $100\%$ & 1(-),3(*) & 14788.32 & 16.79 & $100\%$ & 1(-),2(*) \\
    & -1000 & 11393.85 & 438.19 & $100\%$ & 2(+),3(+)  & 10297.09 & 441.97 & $77\%$ &  1(-),3(*) & 10297.43 & 559.47 & $87\%$ & 1(-),2(*) \\
    \multirow{4}{*}{575-A} & 100 & 29477.29 & 256.00 & $100\%$ & 2(+),3(+) & 28771.41 & 92.60 & $100\%$ & 1(-),3(*) & 28811.27 & 142.93 & $100\%$ & 1(-),2(*)\\
    & 1000 & 18712.86 & 2765.69 & $100\%$ & 2(+),3(+) & -8425.47 & 1464.20 & $100\%$ & 1(-),3(*) & -9557.94 & 1426.20 & $100\%$ & 1(-),2(*)\\
    & -100 & 29488.05 & 83.12 & $100\%$ & 2(+),3(+) & 29369.99 & 50.48 & $100\%$ & 1(-),3(*) & 29352.19 & 35.33 & $100\%$ & 1(-),2(*)\\
    & -1000 & 16297.26 & 1792.34 & $0\%$ & 2(+),3(+) & -9601.48 & 1409.96 & $0\%$ & 1(-),3(*) & -10408.44 & 1201.69 & $0\%$ & 1(-),2(*)\\
    \multirow{4}{*}{1000-A} & 100 & 137195 & 0 & $100\%$ & 2(*),3(-) & 137195 & 0 & $100\%$ & 2(*),3(-) & 143251.53 & 463.14 & $100\%$ & 1(+),2(+)\\
    & 1000 & 137195 & 0 & $100\%$ & 2(*),3(-) & 137195 & 0 & $100\%$ & 2(*),3(-) & 143273.77 & 467.19 & $100\%$ & 1(+),2(+)\\
    & -100 & 137195 & 0 & $100\%$ & 2(*),3(-) & 137195 & 0 & $100\%$ & 2(*),3(-) & 143299.47 & 506.09 & $100\%$ & 1(+),2(+)\\
    & -1000 & 137195 & 0 & $100\%$ & 2(*),3(-) & 137195 & 0 & $100\%$ & 2(*),3(-) & 143253.13 & 436.82 & $100\%$ & 1(+),2(+)\\
    \multirow{4}{*}{51-B} & 100 & 2877.25 & 6.88 & $100\%$ & 2(*),3(-) & 3097.31 & 265.02 & $100\%$ & 1(*),3(-)  & 3614.04 & 128.92 & $100\%$ & 1(+),2(+) \\
    & 1000 & 3822.12 & 217.59 & $100\%$ & 2(*),3(-) & 3771.86 & 239.99 & $100\%$ & 1(*),3(-)  & 3950.61 & 101.59 & $100\%$ & 1(+),2(+)\\ 
    & -100 & 2955.17 & 3.59 & $100\%$ & 2(*),3(-) & 2952.59 & 17.21 & $100\%$ & 1(*),3(-)  & 3198.42 & 78.35 & $100\%$ & 1(+),2(+)\\
    & -1000 & -45.98 & 176.82 & $100\%$ & 2(+),3(*) & -138.90 & 199.22 & $100\%$ & 1(-),3(-) & 54.98 & 137.46 & $100\%$ &  1(*),2(+)\\
    \multirow{4}{*}{225-B} & 100 & 1934.61 & 624.41 & $100\%$ & 2(+),3(+) & 1035.27 & 496.54 & $100\%$ & 1(-),3(*) & 1264.81 & 506.77 & $100\%$ & 1(-),2(*) \\
    & 1000 & 1554.94 & 1166.95 & $100\%$ & 2(+),3(+) & -847.27 & 1726.95 & $100\%$ & 1(-),3(*)  & -1348.48 & 2028.23 & $100\%$ & 1(-),2(*)  \\ 
    & -100 & 1563.13 & 614.52 & $100\%$ & 2(+),3(+) & 994.48 & 529.71 & $100\%$ & 1(-),3(*)  & 1066.07 & 479.07 & $100\%$ & 1(-),2(*) \\
    & -1000 & 228.87 & 451.60 & $100\%$ &  2(+),3(+) & -526.11 & 548.16 & $100\%$ & 1(-),3(*)  & -775.67 & 547.38 & $100\%$ & 1(-),2(*)\\
    \multirow{4}{*}{575-B} & 100 & 302.09 & 1054.91 & $100\%$ & 2(+),3(+) & -7514.01 & 2240.66 & $100\%$ & 1(-),3(*) & -8422.34 & 2271.64 & $100\%$ & 1(-),2(*) \\
    & 1000 & -663.51 & 906.37 & $100\%$ & 2(+),3(+) & -10275.33 & 1339.09 & $100\%$ & 1(-),3(*)  & -10561.94 & 1456.25 & $100\%$ & 1(-),2(*)  \\ 
    & -100 & 1973.34 & 665.31 & $100\%$ & 2(+),3(+) & -2544.96 & 1015.65 & $100\%$ & 1(-),3(*)  & -2869.41 & 1006.13 & $100\%$ & 1(-),2(*) \\
    & -1000 & -8863.53 & 2302.44 & $0\%$ &  2(+),3(+) & -32388.60 & 2188.13 & $0\%$ & 1(-),3(*)  & -31948.07 & 2693.64 & $0\%$ & 1(-),2(*)\\
    \multirow{4}{*}{1000-B} & 100 & 134481 & 0 & $100\%$ & 2(*),3(-) & 134792.97 & 826.06 & $100\%$ & 1(*),3(-) & 142834.43 & 692.54 & $100\%$ & 1(+),2(+) \\
    & 1000 & 134481 & 0 & $100\%$ & 2(*),3(-) & 134583.73 & 272.61 & $100\%$ & 1(*),3(-) & 143007.67 & 692.54 & $100\%$ & 1(+),2(+) \\
    & -100 & 134481 & 0 & $100\%$ & 2(*),3(-) & 134836.07 & 804.90 & $100\%$ & 1(*),3(-)  & 143105.5 & 445.67 & $100\%$ & 1(+),2(+) \\
    & -1000 & 134481 & 0 & $100\%$ & 2(*),3(-) & 134597.57 & 483.35 & $100\%$ & 1(*),3(-)  & 142822.17 & 629.66 & $100\%$ & 1(+),2(+) \\
    51-2 & 100 & 5176.47 & 0 & $100\%$ & 2(+),3(+) & 5175.56 & 3.49 & $100\%$ & 1(-),3(*) & 5175.56 &  3.49 & $100\%$ & 1(-),2(*) \\
    51-u & 100 & 2081.86 & 11.39 & $100\%$ & 2(+),3(+) & 2019.91 & 96.81 & $100\%$ & 1(-),3(*) & 2040.86 & 102.48 & $100\%$ & 1(-),2(*) \\
    51-usw & 100 & 1304.19 & 17.29 & $100\%$ & 2(+),3(+) & 1281.79 & 27.36 & $100\%$ & 1(-),3(*) & 1291.20 & 19.71 & $100\%$ & 1(-),2(*) \\
    150-2 & 100 & 11934.29 & 42.99 & $100\%$ & 2(*),3(+) & 11894.68 & 95.72 & $100\%$ & 1(*),3(*) & 11887.35 & 75.02 & $100\%$ & 1(-),2(*) \\
    150-u & 100 & 7286.82 & 35.59 & $100\%$ & 2(+),3(+) & 7268.77 & 18.39 & $100\%$ & 1(-),3(*) & 7270.01 & 18.31 & $100\%$ & 1(-),2(*) \\
    150-usw & 100 & 3901.91 & 8.34 & $100\%$ & 2(*),3(+) & 3885.13 & 24.21 & $100\%$ & 1(*),3(+) & 3884.37 & 25.83 & $100\%$ & 1(-),2(-) \\
    225-2 & 100 & 21025.80 & 132.98 & $100\%$ & 2(+),3(+) & 20095.64 & 685.48 & $100\%$ & 1(-),3(*) & 19912.64 & 582.11 & $100\%$ & 1(-),2(*) \\
    225-u & 100 & 9069.59 & 74.21 & $100\%$ & 2(+),3(+) & 8207.10 & 430.28 & $100\%$ & 1(-),3(*) & 8259.03 & 469.68 & $100\%$ & 1(-),2(*) \\
    225-usw & 100 & 7244.99 & 290.97 & $100\%$ & 2(+),3(+) & 6910.00 & 278.98 & $100\%$ & 1(-),3(*) & 6855.07 & 251.93 & $100\%$ & 1(-),2(*) \\
    \bottomrule
  \end{tabular}
\end{table}

\begin{table}
    \tiny
    \centering
  \caption{Performance of S4, S5, and C5 with Tour Initialization on TTPTW Benchmark and Their Comparison to DSEA$_1$}
  \label{tab:S4S5Init}
  \begin{tabular}{ccrrcrrcrrcrrc}
    \toprule
    \multirow{2}{*}{Instance} & \multirow{2}{*}{$l$} &  
    \multicolumn{3}{c}{S4-Initial} &
    \multicolumn{3}{c}{S5-Initial} & \multicolumn{3}{c}{C5-Initial} & \multicolumn{3}{c}{DSEA$_1$}\\ \cline{3-14}
    & & mean OB & std OB & FR & mean OB & std OB & FR & mean OB & std OB & FR & mean OB & std OB & FR \\
    \midrule
    \multirow{4}{*}{51-A} & 100 & 1120.40 & 1047.96 & $100\%$ & 308.65 & 0 & $100\%$ & 1323.34 & 1069.58 & $100\%$ & 3834.78 & 28.99 & $100\%$ \\
    & 1000 & 2638.16 & 366.31 & $100\%$ & 498.85 & 0 & $100\%$ & 2876.58 & 305.54 & $100\%$ & 3749.81 & 199.21 & $100\%$ \\
    & -100 &  1948.84 & 0 & $100\%$ &  1948.84 & 0 & $100\%$ &  1948.84 & 0 & $100\%$ & 3468.07 & 0 & $100\%$\\
    & -1000 & -2842.08 & 0 & $0\%$ & -2842.08 & 0 & $0\%$ & -2842.08 & 0 & $0\%$ & -581.81 & 78.99 & $0\%$\\
    \multirow{4}{*}{100-A} & 100 &  4533.98 & 0 & $100\%$ &  4533.98 & 0 & $100\%$&  4533.98 & 0 & $100\%$ & 4533.98 & 0 & $100\%$\\
    & 1000 & 2663.62 & 791.01 & $100\%$  & -1456.52 & 0 & $100\%$ & -1456.52 & 0 & $100\%$ &  4533.98 & 0 & $100\%$\\
    & -100 & 4513.85 & 0 & $100\%$ & 4513.85 & 0 & $100\%$& 4513.85 & 0 & $100\%$ & 4513.85 & 0 & $100\%$ \\
    & -1000 &  4266.41 & 0 & $100\%$ & 3437.13 & 0 & $100\%$ & 3437.13 & 0 & $100\%$ & 4246.19 & 0 & $100\%$ \\
    \multirow{4}{*}{150-A} & 100 &  7629.30 & 0 & $100\%$ & 4749.73 & 0 & $100\%$ & 4749.73 & 0 & $100\%$ & 8772.66 & 12.16 & $100\%$\\
    & 1000 &  945.95 & 0 & $100\%$ & 945.95 & 0 & $100\%$ & 945.95 & 0 & $100\%$ &  8732.84 & 63.46 & $100\%$\\
    & -100 &  8361.17 & 0 & $100\%$ & 3294.41 & 0 & $100\%$ & 3294.41 & 0 & $100\%$  & 8646.24 & 12.16 & $100\%$\\
    & -1000 & 4498.31 & 44.84 & $100\%$ & -3634.12 & 0 & $100\%$ &  -3634.12 & 0 & $100\%$ & 7487.09 & 58.05 & $100\%$\\
    \multirow{4}{*}{225-A} & 100 &  10262.16 & 46.57 & $100\%$ & 3322.41 & 0 & $100\%$ & 3322.41 & 0 & $100\%$ & 14319.98 & 333.14 & $100\%$\\
    & 1000 &  11406.45 & 50.71 & $100\%$ & 8305.97 & 0 & $100\%$ & 8305.97 & 0 & $100\%$ & 11916.29 & 989.30 & $100\%$\\
    & -100 &  12499.20 & 0 & $100\%$ & 189.53 & 0 & $100\%$ & 189.53 & 0 & $100\%$ & 14822.83 & 11.48 & $100\%$\\
    & -1000 & 4637.54 & 749.37 & $0\%$ & -3722.13 & 0 & $0\%$ &  -3722.13 & 0 & $0\%$ &  11393.85 & 438.19 & $100\%$\\
    \bottomrule
  \end{tabular}
\end{table}

\subsection{Impact of Tour Initialization}

Table \ref{tab:S4S5Init} presents the evaluation results of S4, S5, and C5 with the proposed tour initialization algorithm. Compared to Table \ref{tab:S4S5} where the algorithms barely find feasible solutions, the proposed tour initialization algorithm is shown to help them find feasible solutions, except in 51-A and 225-A with $l = -1000$ where the time windows are too tight. They get feasible solutions consistently in most instances because the tour initialization algorithm finds the feasible tour at the beginning. In contrast, when the initialization algorithm cannot find the feasible tour, as in 51-A and 225-A with $l = -1000$, the final solutions are still infeasible. 

Table \ref{tab:S4S5Init} shows that all standard deviations are zero for S5-Initial, where the algorithm starts with Tour Initialization, then repeatedly only generating new solutions using CLK and PackIterative. This is because the CLK fails to find a feasible and better solution than the tour from the initialization.  Similarly, C5-Initial gets the same results, except in 51-A with positive $l$. In these instances, C5-Initial outperforms S4-Initial and S5-Initial. It shows that C5-Initial with a complete pack of the proposed tour initialization, bitflip, insertion, and new tours by CLK can only perform well in small instances. The reason is that it costs too many FEs to carry out many operators, which results in not being able to explore better solutions in larger instances. 

Meanwhile, in bigger instances that have more than $51$ cities, S4-Initial obtains better results than S5-Initial and C5-Initial, although it also shares the same results as the others in 100-A with $l = 100$ and 150-A with $l = 1000$. Thus, the continuation of modifying the initial tour with insertion only (S4-Initial) yields better results than the continuation of generating new tours using CLK and not modifying the initial tour like S5-Initial.

\subsection{Comparison of DSEA with Other Algorithms}

According to the results, DSEA outperforms S4-Init, S5-Init, and C5-Init. Table \ref{tab:S4S5Init} shows that with the same tour initialization algorithm, DSEA$_1$ finds better or at least equal results compared to the other algorithms in almost all instances, except in 100-A with $l = -1000$. It shows that the combination of mutation, Topo and Rain helps DSEA to explore better than Insertion, CLK, and their combination with Bitflip. Furthermore, with tour initialization, all these algorithms do not appear to become increasingly challenged in optimizing solutions as the time windows shrink or the number of cities increases.

Based on the feasible rate in all instances in Table \ref{tab:S4S5} and \ref{tab:S4S5Init}, it is also clear that DSEA is better than S4, S5, C5, LKH-3, and VSR-LKH-3 that find difficulty obtaining feasible solutions. Thus, DSEA outperforms the other algorithms in this study.

\section{Conclusions}\label{concl}
As many real-world problems today involve more than one optimization component and have limited resources, we investigate the Traveling Thief Problem (TTP) which is a combination of the Traveling Salesman Problem (TSP) and the Knapsack Problem (KP) that interdependence to each other.
Moreover, because time is often a critical and limited resource in many real-world applications, this study extends the TTP with time window constraints (TTPTW). We proposed the Dual Search Evolutionary Algorithm (DSEA) for TTPTW. In general, the DSEA algorithms outperform S4, S5, C5, S4-Initial, S5-Initial, C5-Initial, LKH-3, and VSR-LKH-3 to solve TTPTW and perform well in various settings of the time window. Moreover, the proposed tour initialization algorithm helps the algorithms find feasible tours. Based on the experimental result, the addition of the proposed tour initialization to S4, S5 and C5 resulted in a significant improvement compared to S4, S5 and C5 without initialization. DSEA algorithms without packing plan repair (DSEA$_1$) is the best variant among the DSEA variants. This newly defined problem and the benchmark may also be used in future work.

\section*{Acknowledgment}

This study is funded by the Indonesia Endowment Fund for Education (LPDP) of the Ministry of Finance, Republic of Indonesia.


\bibliographystyle{unsrtnat}
\bibliography{references}  






\end{document}